%% file: main.tex
\documentclass[10pt,twocolumn,letterpaper]{article}

\usepackage{iccv}              %

\usepackage[accsupp]{axessibility}  %

\input{preamble}

\definecolor{iccvblue}{rgb}{0.21,0.49,0.74}
\usepackage[pagebackref,breaklinks,colorlinks,allcolors=iccvblue]{hyperref}

\def\paperID{7337} %
\def\confName{ICCV}
\def\confYear{2025}

\newcommand{\methodname}{\textit{MagicDrive-V2}\xspace}
\title{\methodname: High-Resolution Long Video Generation \\ for Autonomous Driving with Adaptive Control}

\newcommand{\urlpath}{\url{https://flymin.github.io/magicdrive-v2/}}
\newcommand{\urlabs}{Project page: {\href{https://flymin.github.io/magicdrive-v2/}{flymin.github.io/magicdrive-v2/}}}

\author{%
  Ruiyuan Gao$^{1}$,
  Kai Chen$^{2}$,
  Bo Xiao$^{3}$,
  Lanqing Hong$^{4}$,
  Zhenguo Li$^{4}$,
  Qiang Xu$^{1}$\\
$^{1}$CUHK
\enspace
$^{2}$HKUST
\enspace
$^{3}$Huawei Cloud
\enspace
$^{4}$Huawei Noah's Ark Lab\\
{\tt\small\{rygao,qxu\}@cse.cuhk.edu.hk},\enspace 
{\tt\small kai.chen@connect.ust.hk}, \\
{\tt\small\{xiaobo15,honglanqing,li.zhenguo\}@huawei.com}
\vspace{-0.3cm}
}

\begin{document}
\maketitle

\input{sec/0_abstract}
\vspace{-0.35cm}
\input{sec/1_intro}
\input{sec/2_related}

\input{sec/3_method}

\input{sec/4_exp}
\input{sec/5_applications}
\input{sec/6_conclusion}

\noindent\textbf{Acknowledgment.}
This work was supported in part by the Innovation and Technology Fund under Grant No. GHP/141/22SZ, in part by the CUHK Strategic Seed Funding for Collaborative Research Scheme under Grant No. 3136023, and in part by the CUHK Research Matching Scheme under Grant No. 7106937, 8601130, and 8601440.
We gratefully acknowledge the support of CANN (Compute Architecture for Neural Networks)
and the Ascend AI Processor used for this research.
We sincerely thank the following individuals for their assistance in conducting human evaluation of generation quality: Dr. Min Li, Dr. Yijun Yang, Mr. Yingji Zhong, Mr. Zhili Liu, Ms. Xiaomeng Wang, Mr. Pengxiang Li, and others.

{
    \small
    \bibliographystyle{ieeenat_fullname}
    \bibliography{main}
}

\input{sec/X_suppl}

\end{document}

%% file: preamble.tex
\usepackage{multirow}

\definecolor{red}{rgb}{0.8,0,0}  
\definecolor{green}{RGB}{0, 133, 21}  
\definecolor{grey}{rgb}{0.5,0.5,0.5}
\definecolor{my-shadow}{HTML}{C0C0C0}

\def\maketitlesupplementary
   {
   \newpage
       {
        \centering
        \Large
        \textbf{\thetitle}\\
        \vspace{0.5em}Supplementary Material (Appendix) \\
        \vspace{1.0em}
       }
   }

\newcommand{\mb}[1]{\mathbf{#1}}

\makeatletter
\def\blfootnote{\xdef\@thefnmark{}\@footnotetext}
\DeclareRobustCommand\onedot{\futurelet\@let@token\@onedot}
\def\@onedot{\ifx\@let@token.\else.\null\fi\xspace}
\def\eg{\textit{e.g}\onedot}

\def\ie{\textit{i.e}\onedot}

\makeatother

\def\tabref#1{Table~\ref{#1}}
\def\figref#1{Figure~\ref{#1}}
\def\secref#1{Section~\ref{#1}}
\def\eqref#1{Equation~\ref{#1}}

\def\appref#1{Appendix~\ref{#1}}

\usepackage{soul}

%% file: sec/0_abstract.tex
\begin{abstract}

The rapid advancement of diffusion models has greatly improved video synthesis, especially in controllable video generation, which is vital for applications like autonomous driving.
Although DiT with 3D VAE has become a standard framework for video generation, it introduces challenges in controllable driving video generation, especially for frame-wise geometric control, rendering existing methods ineffective.
To address these issues, we propose \methodname, a novel approach that integrates the MVDiT block and spatial-temporal conditional encoding to enable multi-view video generation and precise geometric control.
Additionally, we introduce an efficient method for obtaining contextual descriptions for videos to support diverse textual control, along with a progressive training strategy using mixed video data to enhance training efficiency and generalizability.
Consequently, \methodname enables multi-view driving video synthesis with $3.3\times$ resolution and $4\times$ frame count (compared to current SOTA), rich contextual control, and geometric controls.
Extensive experiments demonstrate \methodname's ability, unlocking broader applications in autonomous driving. \urlabs

\end{abstract}

%% file: sec/1_intro.tex
\section{Introduction}
\label{sec:intro}

\begin{figure}[t]
    \centering
    \includegraphics[width=\linewidth]{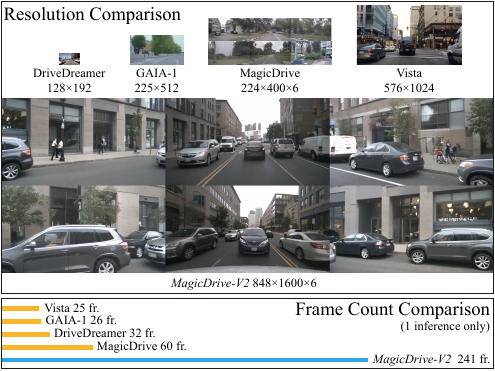}
    \iftoggle{iccvfinal}{
    \vspace{-0.75cm}
    }{
    \vspace{-0.7cm}
    }
    \caption{\methodname generates high-resolution and long videos with multi-view and control supports, significantly exceeding the limitation of previous works~\cite{wang2023drivedreamer,hu2023gaia,magicdrive,gao2024vista}.}
    \label{fig:res_comparison}
    \iftoggle{iccvfinal}{
    \vspace{-0.55cm}
    }{
    \vspace{-0.55cm}
    }
\end{figure}

The rapid advancements in diffusion models and controllable generation has substantially broadened its range of applications~\cite{li2024trackdiffusion,wang2024detdiffusion}.
Especially for autonomous driving, controllable generation of high-resolution and long videos has become a critical research focus~\cite{ma2024unleashing,yang2024drivearena}.
High resolution facilitates the identification of fine details~\cite{liang2022bevfusion}, while long videos provide richer interaction for evaluating and improving algorithms~\cite{yang2024drivearena}.
These capabilities are indispensable for enhancing the performance and reliability of autonomous systems~\cite{magicdrive,wang2023drivedreamer}.
However, due to limitations in model scalability and the compression capability of VAEs~\cite{rombach2021highresolution,blattmann2023stable}, existing methods~\cite{gao2024vista,magicdrive} are significantly constrained by both resolution and frame count, as shown in \figref{fig:res_comparison}.

The introduction of DiT~\cite{peebles2023scalable,opensora} and 3D VAE~\cite{opensora,yang2024cogvideox} has substantially alleviated these challenges. 3D VAE reduces computational overhead by performing spatial-temporal compression, effectively decreasing the computational cost of video generation by an order of magnitude.
DiT enables quality improvements in generation with more model parameters and extensive data.
Consequently, many text-to-video methods leverage DiT with 3D VAE for high-quality generation~\cite{opensora,yang2024cogvideox}.
However, autonomous driving requires multi-view generation with geometry control.
How to effectively leverage the DiT with 3D VAE to achieve such controllable generation remains an open problem.

\begin{figure}
    \centering
    \includegraphics[width=\linewidth]{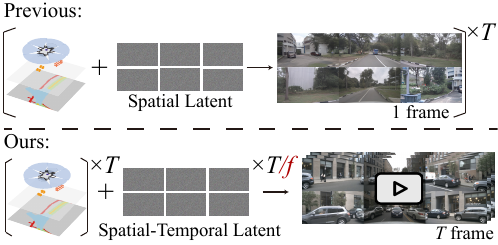}
    \iftoggle{iccvfinal}{
    \vspace{-0.6cm}
    }{
    \vspace{-0.6cm}
    }
    \caption{Different from spatial latent~\cite{magicdrive,wen2024panacea,wang2024driving,ma2024unleashing,jiang2024dive}, spatial-temporal latent requires spatial-temporal condition injection (ours) for frame-wise geometry controls.}
    \label{fig:control}
    \iftoggle{iccvfinal}{
    \vspace{-0.4cm}
    }{
    \vspace{-0.4cm}
    }
\end{figure}

One of the most critical issue from controllable generation in autonomous driving is geometry control, which involves the positioning of objects and the road structures.
These geometric conditions are typically time-dependent, meaning they correspond to individual video frames in a one-to-one manner.
With 2D VAEs~\cite{rombach2021highresolution,blattmann2023stable}, this characteristic allows control methods designed for images~\cite{zhang2023adding} to be seamlessly extended to video generation~\cite{magicdrive}.
However, 3D VAE~\cite{opensora,yang2024cogvideox} disrupts this alignment, as illustrated in \figref{fig:control}.
The spatial-temporal latent of 3D VAE requires that the control conditions also encapsulate corresponding spatial-temporal information.
Therefore, it is necessary to redesign the geometry control module to accommodate the DiT with 3D VAE framework for driving video generation.

In this paper, we present \methodname, a novel approach designed to tackle the challenges of high-resolution and long street-view video synthesis with precise control.
Building upon DiT~\cite{peebles2023scalable,opensora} integrated with 3D VAE~\cite{yang2024cogvideox}, we propose an MVDiT block for multi-view video generation. Additionally, we introduce a novel spatial-temporal conditional encoding for frame-wise control of spatial-temporal latents, enabling precise geometric management.
Additionally, we enhance contextual description supports by efficiently generating captions for each video.
To train the model efficiently, we employ a progressive training strategy, transitioning from short to long videos, which helps the model capture intricate details and generalize to complex scenarios.
Furthermore, we utilize videos of diverse resolutions and durations to enhance the model's generalization capability, enabling it to synthesize longer videos than those used in training.

Our \methodname excels in generating highly realistic videos that align with text, road maps, 3D bounding boxes, and different camera perspectives, achieving significantly higher resolution and more frames than previous works (detailed in \tabref{tab:res-frame}).
Experiments and comparisons demonstrate the effectiveness of our training and control methods, significantly improving controllable street-view video synthesis.
\methodname's flexibility in handling various resolutions, frames, and control signals enables the creation of novel street views suitable for simulations, expanding its potential applications across diverse domains.

In summary, the main contributions of this paper include:
\begin{itemize}
    \item We design an efficient framework, \methodname, leveraging progressive training to achieve high-quality, high-resolution, and long video generation (up to $848\times1600\times6$, 241 frames), significantly superior to prior works.

    \item For geometric controls, novel spatial-temporal control for objects, road semantics, and camera trajectories is proposed while maintaining multi-frame, multi-view consistency. For text control, we enrich the contextual description on driving datasets to support diverse generation.

    \item Our model generalizes well from image to video generation through mixed-resolution and duration training, with extrapolation capabilities, significantly exceeding the resolution and frame number in previous works.
\end{itemize}

%% file: sec/2_related.tex
\begin{figure*}[t]
  \centering
  \includegraphics[width=\linewidth]{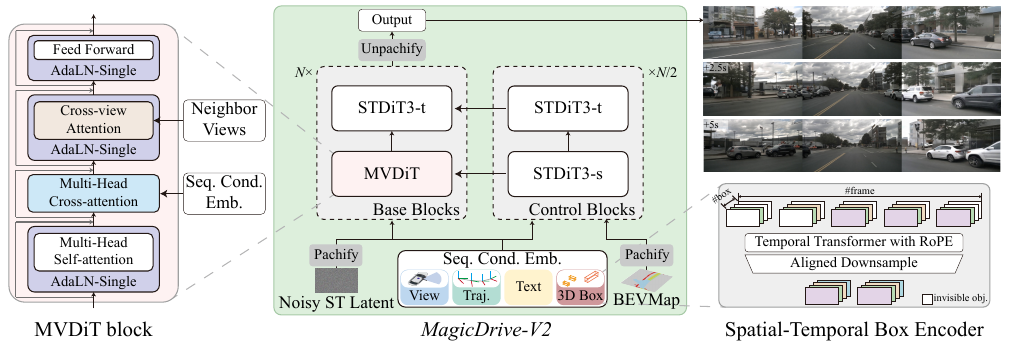}
  \vspace{-0.7cm}
  \caption{\textbf{Architecture Overview of \methodname}. To incorporate different conditions for video generation, \methodname adopts a two-branch architecture as in \cite{zhang2023adding} with basic STDiT3 blocks from \cite{chenpixart,opensora}. We propose the MVDiT block for multi-view consistency and Spatial-Temporal (Box/Traj.) Encoder to inject condition into the Spatial-Temporal (ST) Latent.}
  \label{fig:overview}
  \vspace{-0.3cm}
\end{figure*}

\section{Related Work}
\label{sec:related}

\noindent\textbf{Video Generation in Autonomous Driving.}
Video generation is crucial for autonomous driving, with applications in training perception models~\cite{magicdrive}, testing~\cite{yang2024drivearena}, and scene reconstruction~\cite{gao2024magicdrive3d,zhao2024drivedreamer4d}. It demands extensive fields of view and dynamic object motion handling, requiring precise controllability~\cite{magicdrive} and high-resolution video production~\cite{liang2022bevfusion} with more frames~\cite{gao2024magicdrive3d} and multiple camera perspectives~\cite{zhao2024drivedreamer4d}. Diffusion models have improved controllable multi-view video generation, yet existing models~\cite{wang2024driving,wen2024panacea,zhao2024drivedreamer2,ma2024unleashing,jiang2024dive} lack adequate resolution and frame count for policy testing~\cite{hu2023_uniad} and data engine applications~\cite{magicdrive,gao2024magicdrive3d,zhao2024drivedreamer4d}.
By achieving unprecedented resolution and frame count, as compared in \figref{fig:res_comparison} and \tabref{tab:res-frame}, \methodname addresses the limitations of previous driving video generation methods, enabling more potential applications.

\input{tables/comparison}

\noindent\textbf{Diffusion Models and DiT Architectures.}
Diffusion models~\citep{ho2020denoising,song2020score,zheng2023non} generate data by learning the denoising steps from Gaussian noises to samples, widely used in image~\cite{dhariwal2021diffusion,rombach2021highresolution,wang2024detdiffusion} and video~\cite{ho2022video} generation.
From the modeling's perspective, Flow Matching~\cite{lipmanflow,esser2024scaling} simplifies diffusion models, enhancing the efficiency of training and inference.
From the implementation's perspective, the architecture shifts from UNet~\cite{ho2020denoising} to DiT, due to better scalability~\cite{peebles2023scalable,chenpixart}, especially for high-resolution tasks~\cite{chen2024pixart}. Our \methodname also leverages Flow Matching and DiT for scaling to high-resolution and long video generation.

\noindent\textbf{Conditional Generation.}
Conditional generation is crucial for various applications utilizing generative models. Cross-attention layers from LDM~\cite{rombach2021highresolution} and ControlNet's~\cite{zhang2023adding} additive encoding for grid-shaped control signals are leading methods in controllable diffusion-based generation. In the street-view generation, MagicDrive~\cite{magicdrive} and MagicDrive3D~\cite{gao2024magicdrive3d} integrate 3D bounding boxes, BEV maps, ego trajectories, and camera poses for multi-view street scene synthesis. However, these methods are limited to spatial encoding and are not directly applicable to spatial-temporal latents from 3D VAE~\cite{yang2024cogvideox}, as shown in \figref{fig:control}.
\methodname presents a novel control paradigm for spatial-temporally compressed latents from 3D VAE, enabling the controllable generation of high-resolution long videos.

%% file: tables/comparison.tex
\begin{table}[t]
\centering
\begin{tabular}{@{}clrr@{}}
\toprule
Type & Method & Total Res. & Frame \\ \midrule
\multirow{3}{*}{\begin{tabular}[c]{@{}c@{}}Front\\ View\end{tabular}} & GAIA-1$^*$\cite{hu2023gaia} & 288$\times$512$\times$1 & 26\\
 & DriveDreamer~\cite{wang2023drivedreamer} & 128$\times$192$\times$1 & 32\\
 & Vista$^*$~\cite{gao2024vista} & 576$\times$1024$\times$1 & 25 \\ \midrule
\multirow{9}{*}{\begin{tabular}[c]{@{}c@{}}Multi-\\ view\end{tabular}}
 & MagicDrive~\cite{magicdrive} & 224$\times$400$\times$6 & 60 \\
 & Drive-WM~\cite{wang2024driving} & 192$\times$384$\times$6 & 8\\
 & Panacea~\cite{wen2024panacea} & 256$\times$512$\times$6 & 8\\
 & DriveDreamer2~\cite{zhao2024drivedreamer2} & 256$\times$448$\times$6 & 8 \\
 & Delphi~\cite{ma2024unleashing} & 512$\times$512$\times$6 & 10 \\
 & DiVE~\cite{jiang2024dive} & 480p$\times$6 & 16 \\
 \cmidrule{2-4}
& \methodname & \textbf{848$\times$1600$\times$6} & \textbf{241} \\
\bottomrule
\end{tabular}
\iftoggle{iccvfinal}{
\vspace{-0.3cm}
}{
\vspace{-0.2cm}
}
\caption{\textbf{Comparison of Resolution and Frame Count}. We only consider a single inference, since rollout notably degrades quality (see \appref{app:vsrollout}). $^*$Only support text \& image(s) conditions.}
\label{tab:res-frame}
\iftoggle{iccvfinal}{
\vspace{-0.5cm}
}{
\vspace{-0.4cm}
}
\end{table}

%% file: sec/3_method.tex
\section{Preliminary}\label{sec:pre}
\noindent\textbf{Problem Formulation}.
This paper addresses controllable high-resolution and long video generation for street views. Given a sequence of frame descriptions \(\{\mb{S}_{t}\}, t \in \{0, \dots, T\}\), the goal is to generate corresponding street-view videos from latent variables \(\mb{z} \sim \mathcal{N}(\mb{0}, \boldsymbol{I})\), i.e., \(\{\mb{I}_{c,t}\} = \mathcal{G}(\{\mb{S}_{t}\}, \mb{z})\), where \(c \in \{0, \dots, C\}\) denotes the \(C\) camera views. Here, high resolution indicates that \(\mb{I}\) has a high resolution, and long video implies that \(T\) is large.

To describe a street-view video, we adopt the conditions outlined in \cite{gao2024magicdrive3d,magicdrive}. Specifically, frame descriptions \(\mb{S}_{t}=\{\mb{C}, \mb{M}_{t}, \mb{B}_{t}, \mb{L}, \mb{Tr}_{t}^{0}\}\) include the camera poses \(\{\mb{C}_{c}\}=[\mb{R}_{c},\mb{t}_{c}]\)\footnote{Typically, cameras are fixed throughout the video.}, a road map \(\mb{M}_{t}\in\{0, 1\}^{w\times h\times c}\) representing a \(w\times h\) meter road area in BEV with \(c\) semantic classes, 3D bounding boxes \(\mb{B}_{t}=\{(c_i, b_i)\}_{i=1}^{N}\) where each object is described by a box \(b_i=\{(x_j, y_j, z_j)\}_{j=1}^{8}\in\mathbb{R}^{8\times 3}\) and class \(c_i\in\mathcal{C}\), text \(\mb{L}\) adding information for the whole video (e.g., weather and time of day), and the ego vehicle trajectory \(\mb{Tr}_{t}^{0}\) which describes the transformation from the LiDAR coordinate of each frame to the first frame, i.e., \(\mb{Tr}^{0}_{t}=[\mb{R}_{t}^{0},\mb{t}_{t}^{0}]\). 
All geometric data, except \(\mb{Tr}_{t}^{0}\), is parameterized in the ego car's local LiDAR frame.

\noindent\textbf{LDMs and Flow Matching}.
For high-resolution image generation, \citet{rombach2021highresolution} propose Latent Diffusion Models (LDMs), using a pre-trained VAE for image downsampling and diffusion models for latent generation.
This manner is widely adopted in both image generation~\cite{chenpixart,esser2024scaling} and video generation~\cite{blattmann2023stable,opensora,videoworldsimulators2024}.
\methodname is also based on the VAE+diffusion formulation.
With the latest advancement on diffusion models, \citet{esser2024scaling} propose to train large scale diffusion models through simulation-free rectified flow~\cite{liuflow,lipman2023flow} and v-prediction loss~\cite{esser2024scaling}:
\setlength{\abovedisplayskip}{0.5em}
\setlength{\belowdisplayskip}{0.5em}
\begin{align}
    \mb{z}_{t} &= t \mb{z}_{1} + (1-t)\epsilon\\
    \mathcal{L}_{CFM} &= \mathbb{E}_{\epsilon\sim \mathcal{N}(0,I)}\|v_{\Theta}(\mb{z}_{t},t)-(\mb{z}_{1}-\epsilon)\|_{2}^{2}\text{,}
\end{align}
where $\mb{z}_{1}$ is the latent of video from the VAE, $t\sim\operatorname{lognorm}(0,1)$ is timestep and $v_{\Theta}$ is the model.

\section{Methods}\label{sec:method}

\subsection{Overview of \methodname}
\figref{fig:overview} illustrates an overview of the model architecture.
Based on the STDiT-3 blocks~\citet{opensora}, we introduce Multi-View DiT (MVDiT) block for multi-view generation and employ cross-attention~\cite{rombach2021highresolution} for text, boxes, camera views, and trajectories, alongside an addictive branch~\cite{zhang2023adding} for maps to inject control signals.

Nevertheless, the spatial encoding of control signals~\cite{magicdrive,ma2024unleashing,wen2024panacea,wang2024driving,zhao2024drivedreamer2} is incompatible with spatial-temporal latents, as illustrated in \figref{fig:control}. Consequently, the encoder for each control signal is reformulated, exemplified by the spatial-temporal box encoder on the right of \figref{fig:overview}. Further details are provided in \secref{sec:control}. 

Besides, \methodname enriches text control beyond previous works (\eg.~\cite{magicdrive}) by generating contextual description through image captioning (\secref{sec:text control}).
Furthermore, \methodname employs progressive training from images to high-resolution short videos, then to long videos, enhancing diffusion model convergence. Using variable-length, multi-resolution videos in training also enables image and video generation at diverse resolutions and frame extrapolation beyond training settings (\secref{sec:bootstrap-variable}).

\subsection{Architecture for High-Resolution Long Video}\label{sec:design}
Training high-resolution, long-video diffusion models is computationally intensive, requiring significant GPU memory.
DiT and 3D VAE are key to scaling these models. As noted by \citet{peebles2023scalable}, DiT's lower Gflops correlate with better FID, making it more efficient than UNet~\cite{magicdrive}.
For memory, 3D VAE~\cite{yang2024cogvideox} compresses temporal information by $4\times$, achieving a $256\times$ compression ratio\footnote{We ignore latent dimension discrepancies.}, reducing sequence length and memory use, especially for transformers~\cite{dao2022flashattention}.
Architectural unification~\cite{sun2024autoregressive} enables DiT to leverage advanced parallelization, \eg, sequence parallelism~\cite{opensora}, overcoming single GPU memory limits for higher resolution and longer videos.

As depicted in \figref{fig:overview}, \methodname utilizes STDiT-3 blocks as per \citet{opensora}.
The architecture design incorporates two significant modifications.
First, we propose a Multi-View DiT (MVDiT) block, integrating a cross-view attention layer~\cite{magicdrive} to facilitate multi-view generation, as shown on the left of \figref{fig:overview}.
Second, to handle multiple geometric controls, \methodname utilizes cross-attention~\cite{rombach2021highresolution} for processing text, boxes, camera views, and trajectories, while incorporating an additive branch~\cite{zhang2023adding} for maps to inject road structure controls.

\subsection{Spatial-Temporal Control on 3D VAE}\label{sec:control}
A major challenge with 3D VAE adoption is geometric control. As in \figref{fig:control}, geometric control manages per-frame content spatially. 
Previously with 2D VAE, the temporal axis persists in geometric descriptors $\{\mb{S}_{t}\}, t\in\{1,...,T\}$, video latent and the decoded video.
This temporally aligning feature enables video geometric control with spatial control methods similar to that on images~\cite{meng2022sdedit}.
However, 3D VAE yields $T/f$ latents (where $f$ is the temporal compression ratio), misaligning control signals with latents and rendering previous control techniques~\cite{magicdrive,wen2024panacea} ineffective.

Besides, the issue of temporal misalignment does not exist in ambiguous control conditions (\eg, text), as these conditions are typically employed to describe the entire video.
Specifically, the temporal dimension of such conditions is $1$, and they are repeated for $T/f$ times to control the entire video.
Consequently, a naive approach to handle such dimension misalignment would be to follow the method of text control, where the temporal dimension is first reduced to $1$ through a temporal encoder and then repeated to $T/f$ for control.
However, our experiments reveal that this operation results in trailing issues (as "Box Reduce" in \figref{fig:abl_vis}), which we hypothesize are caused by the repeat operation.
Therefore, we design a downsampling method that aligns the dimensions while preserving the distinctiveness of temporal information, thereby enabling precise control of geometric information for each frame.

\begin{figure}[t]
    \centering
    \includegraphics[width=0.95\linewidth]{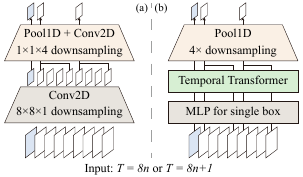}
    \vspace{-0.3cm}
    \caption{\textbf{Spatial-Temporal Encoder for Maps} (a) \textbf{and Boxes} (b). Our spatial encoding module follows \cite{magicdrive}, and temporal encoding integrates the downsampling strategy used in our 3D VAE~\cite{yang2024cogvideox}, resulting in temporally aligned embedding between the control signals and video latents.}
    \label{fig:downsample}
    \vspace{-0.3cm}
\end{figure}

Our spatial-temporal encoding modules are illustrated in \figref{fig:downsample}.
This involves re-aligning the maps ($\mb{M}_{t}$), boxes ($\mb{B}_{t}$), and trajectory ($\mb{Tr}_{t}^{0}$) within the scene descriptor ($\mb{S}_{t}$). Maps, represented as grid data, are straightforward to manage. By extending ControlNet's~\cite{zhang2023adding} design, we utilize temporal downsampling modules with new trainable parameters in 3D VAE to align features between control and base blocks, as depicted in \figref{fig:downsample}(a).

For 3D boxes, padding is applied to invisible boxes to maintain consistent box sequence length across views and frames, illustrated in \figref{fig:overview} right.
We employ a downsampling module with a temporal transformer and RoPE~\cite{su2024roformer} to capture temporal correlation, creating spatial-temporal embeddings aligned with video latents, as in \figref{fig:downsample}(b). The spatial-temporal encoder for boxes can also be adapted for ego trajectory ($\mb{Tr}_{t}^{0}$), by replacing the MLP for boxes with an MLP for camera poses~\cite{magicdrive}.

All the downsampling ratios are aligned with the adopted 3D VAE~\cite{yang2024cogvideox}, \ie, with $8n$ or $8n+1$ as inputs, and $2n$ or $2n+1$ as outputs, respectively.
\secref{sec:ablation} holds more ablation studies on different downsampling methods.

\begin{figure}
    \centering
    \includegraphics[width=\linewidth]{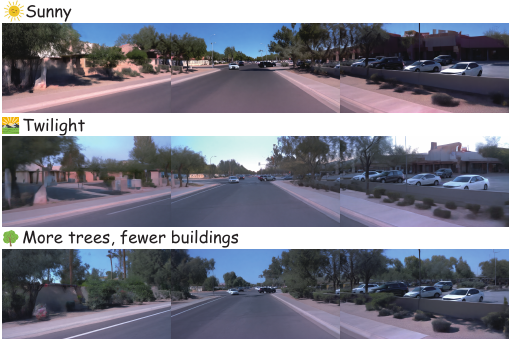}
    \vspace{-0.7cm}
    \caption{\methodname supports more diverse text control by enriching the textual description of driving datasets (\eg, nuScene~\cite{nuScenes}, Waymo~\cite{Sun_2020_CVPR}) than previous works (\eg, \cite{magicdrive}). Please see the full videos on our project page.}
    \label{fig:diverse}
    \vspace{-0.4cm}
\end{figure}

\subsection{Enrich Text Control by Image Caption}\label{sec:text control}
Although geometric descriptors offer precise control over objects and maps, textual descriptions in autonomous driving datasets (\eg, nuScenes) are often restricted to basic weather (sunny/rain) and time (day/night), lacking richer scene context such as road types (\eg, highway/urban) or background elements (\eg, buildings, trees).
To enhance textual control, we regenerated video captions using a Multimodal Large Language Model (MLLM)~\cite{hurst2024gpt}.
By prompting MLLM to focus on aspects beyond object categories, trajectories, and road structures, we ensure efficiency and consistency.
Specifically, first, it eliminates the need for the MLLM to describe dynamics, thus only the middle frame of each video is used; second, it avoids potential conflicts between multiple control signals.
\figref{fig:diverse} shows the diverse contextual description supports of \methodname.

\subsection{Progressive Training with Mixed Video Types}\label{sec:bootstrap-variable}
To accelerate model convergence, we employ a three-stage training strategy: starting with low-resolution images, transitioning to high-resolution short videos, and finally using high-resolution long videos.
This approach is based on two key insights.
First, in controllable generation, models prioritize content quality before learning controllability, also observed by \citet{gao2024vista}. Our progressive training enables faster controllability acquisition.
Second, models adapt more quickly to high resolution than to long videos, so we emphasize short-video training early on.
Details of the mixing method are provided in the \appref{app:mix}.

In the third stage, we mix videos up to 241 frames (the maximum frame count of the dataset) at 224$\times$400 and a resolution of up to 848$\times$1600 (the maximum resolution of nuScenes~\cite{nuScenes}) at 33 frames.
This mixed-resolution and mixed-length training allows the model to generalize across dimensions, supporting flexible generation and extrapolation beyond training configurations (e.g., up to 848×1600 at 241 frames as demonstrated in \secref{sec:application}).
Ablation studies in \secref{sec:ablation} further validate this approach.

%% file: sec/4_exp.tex
\section{Experiments}
\label{sec:exp}

\subsection{Experimental Setups}
\noindent\textbf{Dataset and Baselines.}
We employ the nuScenes dataset~\citep{nuScenes}, a prominent dataset for street-view generation~\cite{magicdrive,yang2023bevcontrol,ma2024unleashing}, to evaluate \methodname.
We adhere to the official splits, using 700 multi-view videos for training and 150 for validation. Our primary baseline is MagicDrive~\cite{magicdrive}, comprising three models: MagicDrive (16f) for 16-frame video generation; MagicDrive (60f), which extends MagicDrive (16f) to 60 frames, both from \citet{magicdrive}; and MagicDrive3D, a 16-frame model from \citet{gao2024magicdrive3d}.
Moreover, we employed the Waymo Open Dataset~\cite{Sun_2020_CVPR} to further fine-tune the model, aiming to validate its scalability and to facilitate the generation of videos with diverse environmental styles and different numbers of perspectives.
Further details are provided in \appref{app:exp_setup}. 

\input{tables/video_eval}

\begin{figure*}
    \centering
    \includegraphics[width=0.98\linewidth]{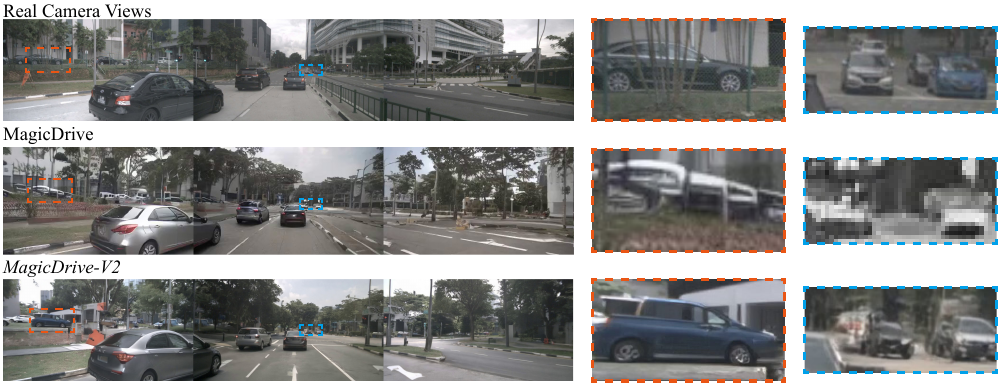}
    \vspace{-0.3cm}
    \caption{\textbf{Qualitative Comparison between \methodname and MagicDrive~\cite{magicdrive}}.
    Frames are extracted from the generated videos.
    To conserve space, we only present the 3 (out of 6) perspectives that include the front view.
    Two crops from the generated views are magnified and shown on the right.   
    By generating 4$\times$ resolution of MagicDrive,
    the synthesized street view of our model contains finer details.}
    \label{fig:comparison}
    \vspace{-0.4cm}
\end{figure*}

\noindent\textbf{Evaluation Metrics.}
We evaluate both realism and controllability in street-view generation for videos and images. For video generation, we follow benchmarks from \cite{w-coda}, utilizing FVD for video quality; mAP from 3D object detection and mIoU from BEV segmentation for controllability, using BEVFormer~\cite{li2022bevformer} for both tasks, which is a video-based perception model.
For image generation, we adopt the metrics from \citet{magicdrive}, employing FID for image quality; mAP with BEVFusion~\citep{liu2022bevfusion} and road mIoU with CVT~\citep{zhou2022cross} for controllability assessment. BEVFusion and CVT are image-based models.

\noindent\textbf{Model Setup.}
In accordance with \secref{sec:design}, we adopt the 3D VAE framework from CogVideoX~\cite{yang2024cogvideox} and train diffusion models for street-view video generation from scratch\footnote{For more discussion on this choice, see \secref{sec:ablation} and \appref{app:vae}.}. Initially, the model comprises only spatial blocks for base and control blocks, focusing on image generation.
Temporal blocks are incorporated in the second stage, forming the full architecture of \methodname.
Further training details can be found in \appref{app:exp_train}.

\subsection{Results and Analysis}\label{sec:results}

\noindent\textbf{Generation Quality}.
\methodname excels in both video and image generation tasks. 
In video tasks, it significantly reduces FVD compared to MagicDrive (\tabref{tab:video}), due to the DiT architecture enhancing inter-frame consistency and spatial-temporal condition encoding for precise control of object motion and positioning.
Further illustrated in \figref{fig:comparison}, \methodname generates high-resolution videos that not only improve quality but also incorporate more intricate details, closely resembling footage captured by real cameras.
This enhancement is achieved through our advanced training on variable lengths and resolutions, allowing for more realistic and detailed outputs.
We further conduct human evaluation on consistency in \appref{app:human-eval}.

Beneficial from the mixing training approach, \methodname is capable of image generation.
As shown in \tabref{tab:image}, \methodname matches baseline performance in multi-view street-view tasks and surpasses baselines in vehicle segmentation mIoU and object detection mAP. This demonstrates the strong generalization capabilities of our spatial-temporal condition encoding.

\begin{figure*}
    \centering
    \includegraphics[width=\linewidth]{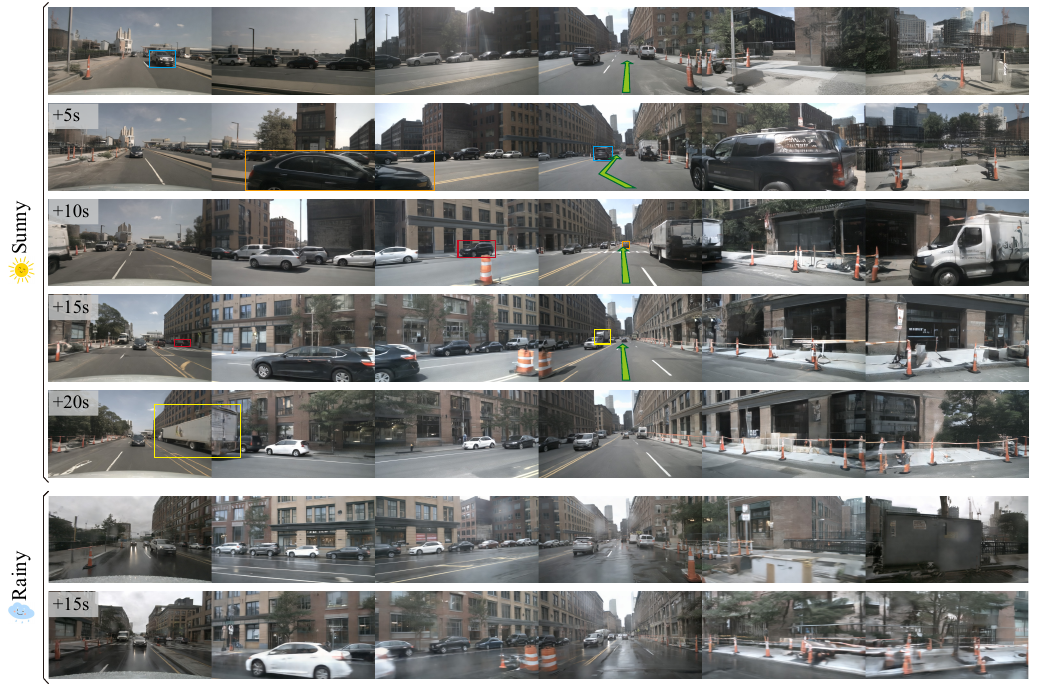}
    \vspace{-0.7cm}
    \caption{\methodname generates high-resolution (e.g., 424$\times$800 here) street-view videos for 241 frames (i.e., the full length of nuScenes videos, approximately 20 seconds at 12 FPS) with multiple controls (i.e., road map, object boxes, ego trajectory, and text). Notably, the 241-frame length at 424$\times$800 is unseen during training, demonstrating our method's generalization capability to video length. We annotate the ego-vehicle trajectory and selected objects to aid localization, with same-color boxes denoting the same object. Due to space constraints, the ``rainy'' example includes only two frames; additional examples can be found in \appref{app:more vis} and on our project page.}
    \label{fig:main}
    \vspace{-0.4cm}
\end{figure*}

\noindent\textbf{Controllability}.
The quantitative results presented in \tabref{tab:video} and \ref{tab:image} demonstrate that the images and videos generated by \methodname effectively reflect the control conditions. Additionally, \figref{fig:diverse} and \ref{fig:main} provide visual results, showing that multiple control conditions can independently influence the generated content.
For instance, weather can be altered via text input (from sunny to rainy) while maintaining road structure, and the trajectories of other vehicles and the ego vehicle.
By varying the combinations of conditions, \methodname is capable of producing a diverse range of high-quality street-view videos.

\input{tables/image_eval}

\begin{figure}[t]
    \centering
    \includegraphics[width=0.95\linewidth]{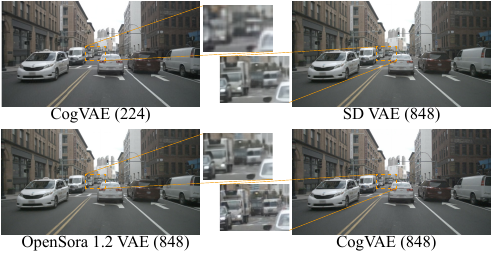}
    \vspace{-0.3cm}
    \caption{
    \textbf{Reconstruction Visualization from Different VAEs}.
    CogVAE~\cite{yang2024cogvideox} maintains most details compared with others and exhibits better performance for high-resolution content.
    }
    \label{fig:vae}
    \vspace{-0.5cm}
\end{figure}

\subsection{Ablation Studies}\label{sec:ablation}

\noindent\textbf{VAE Comparison for Street Views}.
Before training the diffusion models, we evaluate the performance of open-source 3D VAEs (\ie, CogVAE~\cite{yang2024cogvideox} and Open-Sora~\cite{opensora}) on street views, in comparison with the 2D SD VAE~\cite{rombach2021highresolution}.
As illustrated in \figref{fig:vae}, CogVAE consistently outperforms its counterparts on reconstruction ability.
Besides, as presented in \appref{app:vae_comp}, CogVAE demonstrates minimal performance degradation over longer video sequences, making it particularly well-suited for long video generation tasks.
Additionally, we found that all VAEs exhibit improved reconstruction capabilities with increasing resolution.
This insight is beneficial for advancing our model's ability to generate high-quality images and videos by focusing on higher-resolution outputs.
For more detailed quantitative results and discussion, see \appref{app:vae_comp}.

\begin{figure}[t]
    \centering
    \includegraphics[width=\linewidth]{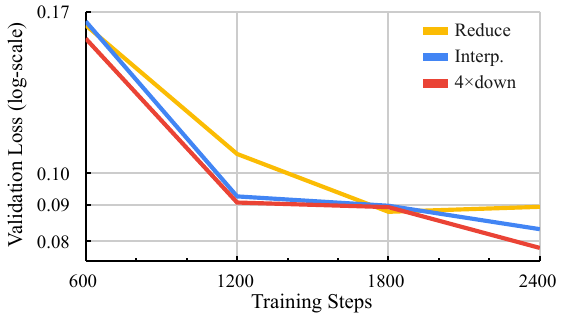}
    \vspace{-0.8cm}
    \caption{\textbf{Validation Loss through Training with Different ST Encodings}.
    $4\times$down (our methods in \methodname) can help the model converge, performing the best among all the encodings.}
    \label{fig:abl_down}
    \vspace{-0.6cm}
\end{figure}

\begin{figure*}[t]
    \centering
    \includegraphics[width=0.95\linewidth]{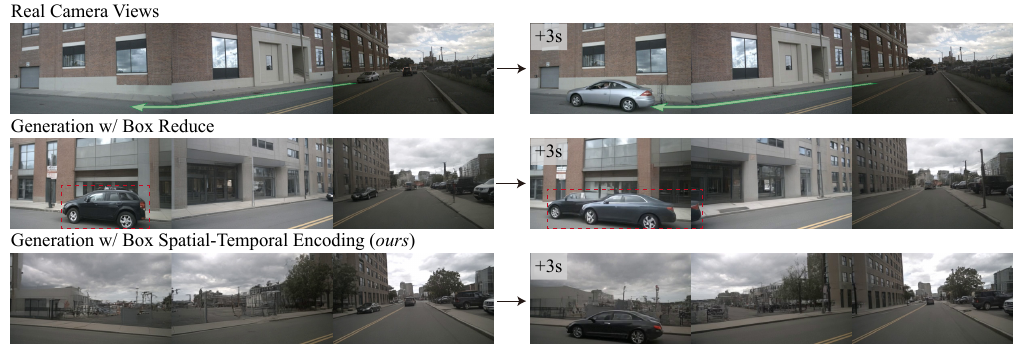}
    \vspace{-0.3cm}
    \caption{
    \textbf{Visual Effect of Spatial-Temporal Encoding for Boxes}.
    For videos with spatial-temporal latent encoding, a simple global reduce baseline can cause artifacts and trailing effects in object trajectories across viewpoints (we highlight them with {\color{red}red boxes}). Our spatial-temporal encoding effectively resolves this, maintaining object clarity and accurate motion trajectories.
    }
    \label{fig:abl_vis}
    \vspace{-0.3cm}
\end{figure*}

\noindent\textbf{Spatial-Temporal Conditioning}.
We demonstrate the spatial-temporal encoder's effectiveness by the validation loss in over-fitting experiments (\figref{fig:abl_down}) and visualization comparison (\figref{fig:abl_vis}).
We compare two baselines: global temporal dimension reduction (Reduce) and temporal dimension interpolation (Interp.) for alignment. In over-fitting training with 16 samples, the $4\times$ downsampling technique ($4\times$ down, ours) accelerates convergence and achieves the lowest final validation loss~\cite{esser2024scaling}, as shown in \figref{fig:abl_down}. Furthermore, \figref{fig:abl_vis} illustrates that, unlike the global reduction baseline, $4\times$ down reduces artifacts and maintains accurate motion trajectories. These results confirm the spatial-temporal encoder's capability to enhance data encoding and improve video generation performance using spatial-temporal latents.

\noindent\textbf{Variable Length and Resolution Training}. As presented in \secref{sec:bootstrap-variable}, \methodname adopts a training strategy by mixing different length and resolution videos. Our ablation study demonstrates the effectiveness of this strategy.
As shown in \tabref{tab:mixing}, the limitations of VAE (as discussed in \secref{sec:results}) are evident when using only low-resolution videos (17$\times$224$\times$400), indicated by worse FVD and lower mAP and mIoU scores compared to other settings.
Incorporating longer videos improves the model's controllability (both mAP and mIoU are higher), while incorporating high-resolution videos enhances overall quality (all three metrics are notably improved).

Although mixing different frame lengths slightly degrades FVD, it is crucial for enabling the model to generate videos of various lengths and extrapolate to unseen lengths (see \secref{sec:application}).
Thus, we combine both resolutions and frame lengths, effectively balancing the trade-offs between video quality, controllability, and model functionality.

\input{tables/mixing}

%% file: tables/video_eval.tex
\begin{table}[t]
\centering
\begin{tabular}{l|ccc}
\toprule
Method & FVD$\downarrow$ & mAP$\uparrow$ & mIoU$\uparrow$ \\ \midrule
MagicDrive~\cite{magicdrive} (16f) & 218.12 & 11.86 & 18.34 \\
MagicDrive~\cite{magicdrive} (60f) & 217.94 & 11.49 & 18.27 \\
MagicDrive3D~\cite{gao2024magicdrive3d} & 210.40 & 12.05 & 18.27 \\ \midrule
\methodname
    & \textbf{94.84} & \textbf{18.17} & \textbf{20.40} \\
\bottomrule
\end{tabular}
\vspace{-0.2cm}
\caption{\textbf{Comparison with Baselines for Controllable Video Generation}.
Videos are generated according to conditions from the nuScenes validation set. Only first 16 frames are kept for evaluation, as in \cite{w-coda}. 
$\uparrow$/$\downarrow$ indicates that a higher/lower value is better.
}
\label{tab:video}
\vspace{-0.3cm}
\end{table}

%% file: tables/image_eval.tex
\begin{table}[t]
\centering
\setlength{\tabcolsep}{4pt}
\begin{tabular}{l|c|cc|c}
\toprule
Method & \begin{tabular}[c]{@{}c@{}}FID\\$\downarrow$\end{tabular} & \begin{tabular}[c]{@{}c@{}}Road\\mIoU$\uparrow$\end{tabular} & \begin{tabular}[c]{@{}c@{}}Vehicle\\mIoU$\uparrow$\end{tabular} & \begin{tabular}[c]{@{}c@{}}mAP\\$\uparrow$\end{tabular} \\ \midrule
BEVControl~\cite{yang2023bevcontrol}
    & 24.85
    & 60.80
    & 26.80
    & N/A \\ 
MagicDrive~\cite{magicdrive} (Img) & \textbf{16.20} & \textbf{61.05} & 27.01 & 12.30 \\ \midrule
\methodname 
&  20.91	& 59.79	& \textbf{32.73} & \textbf{17.65} \\ \bottomrule
\end{tabular}
\vspace{-0.35cm}
\caption{
\textbf{Comparison with Baselines for Controllable Image Generation}.
All the annotations \& camera views from the nuScenes validation set are used for evaluation. $\uparrow$/$\downarrow$ indicates that a higher/lower value is better.
}
\label{tab:image}
\vspace{-0.45cm}
\end{table}

%% file: tables/mixing.tex
\begin{table}[t]
\centering
\begin{tabular}{@{}l|ccc@{}}
\toprule
Training Data & FVD$\downarrow$ & mAP$\uparrow$ & mIoU$\uparrow$ \\ \midrule
17$\times$224$\times$400
& 97.21 & 10.17 & 12.42 \\
(1-65)-224$\times$400
& 100.73 & 10.51 & 12.74 \\
17$\times$(224$\times$440 - 424$\times$800)
& \textbf{96.34} & 14.91 & 17.53 \\ %
\midrule
1-65$\times$(224$\times$440 - 424$\times$800) & 99.66 & \textbf{15.44} & \textbf{18.26} \\ \bottomrule
\end{tabular}
\vspace{-0.3cm}
\caption{
\textbf{Comparison between Different Training Configurations}. To test adaptation ability for higher resolution and longer videos, all the models load a pre-trained weight for short videos (9$\times$424$\times$800) and are trained with the same GPU hours.
} 
\label{tab:mixing}
\vspace{-0.5cm}
\end{table}

%% file: sec/5_applications.tex
\section{Applications}\label{sec:application}

\noindent\textbf{Extrapolation for Longer Video Generation}.
Through variable length and resolution training (\secref{sec:bootstrap-variable}), \methodname effectively generates videos exceeding the training setup's length. Though trained on videos up to 33$\times$848$\times$1600 and 241$\times$224$\times$400, \methodname successfully produces 241$\times$848$\times$1600 videos (full length and full resolution of nuScenes), which are $8\times$ the length of the training samples, as demonstrated in \figref{fig:main} and videos on our project page.
To further validate this capability, we compare the first-16-frame FVD for short videos (17 frames) and per-16-frame FVD for longer videos (65 frames and above). As shown in \tabref{tab:long fvd}, the 16-frame FVDs remain consistent across both seen and extrapolated configurations, extending to 129$\times$848$\times$1600$\times$6 generation (this is not the upper limit; please see note in the Appendix).
This affirms the model's robust generalization capability. 

\input{tables/long_fvd}

\noindent\textbf{Fast Generalization on Other Datasets}.
To evaluate the generalization of \methodname, we fine-tuned the Stage 3 model on the Waymo dataset~\cite{Sun_2020_CVPR}, enabling rapid generation of 3-view videos in $1$ day ($1$k+ steps) with strong controllability (examples shown in \figref{fig:diverse} and more videos can be found on our project page).
Moreover, mixed training on Waymo and nuScenes datasets further enhanced the model’s ability to generate videos with varying perspectives, improving overall quality. The final model achieved an FVD of 105.17 on Waymo and 74.30 on nuScenes, surpassing results in \tabref{tab:video}.

%% file: tables/long_fvd.tex
\begin{table}[t]
\centering
\setlength{\tabcolsep}{4pt}
\begin{tabular}{ccccc}
\toprule
\multirow{2.5}{*}{Resolution} & First-16- & \multicolumn{2}{c}{Avg. of Per-16-Frame} \\ \cmidrule{3-5} 
 & Frame & 2$\times$ & 3$\times$ & 4$\times$ \\ \midrule
424$\times$800 & 530.65 & 562.99 & / & / \\\midrule
848$\times$1600 & 559.70 & 573.46 & 583.50 & 585.89 \\ \bottomrule
\end{tabular}
\vspace{-0.3cm}
\caption{
\textbf{Generation Quality for Videos Longer than Training}.
We randomly sample 10 sequences from the nuScenes validation set and report FVD (the lower the better). $n\times$: $n$ times of maximum training frame number, \ie, 129 frames for 424$\times$800 and 33 for 848$\times$1600. /: exceed the maximum frame of dataset.
}
\label{tab:long fvd}
\vspace{-0.5cm}
\end{table}

%% file: sec/6_conclusion.tex
\section{Conclusion and Discussion}\label{sec:discussion}
In this paper, we presented \methodname, an innovative framework for high-resolution and long video synthesis with precise control, specifically tailored for applications like autonomous driving. By introducing MVDiT block and spatial-temporal conditional encoding module, \methodname effectively addresses the challenges of scalability and geometry control in video generation with the DiT architecture and 3D VAE.
Our approach also introduces richer contextual description for driving video generation, and a progressive training strategy with variable length and resolution adaption, enhancing the model's generation ability.
Extensive experiments demonstrate that \methodname generates realistic videos that maintain spatial and temporal coherence, significantly surpassing previous methods in resolution 
and frame count, opening new possibilities for simulations and other applications in autonomous driving.

%% file: sec/X_suppl.tex
\clearpage
\onecolumn
\setcounter{page}{1}
\setcounter{section}{0}
\renewcommand{\thesection}{\Alph{section}}
\setcounter{figure}{0}
\renewcommand{\thefigure}{\Roman{figure}}
\setcounter{table}{0}
\renewcommand{\thetable}{\Roman{table}}
\maketitlesupplementary

\begin{center}
\noindent Please find the videos on our project website: \urlpath   
\end{center}

\noindent\textbf{Note}: Our model is capable of generating videos at a resolution of 848$\times$1600 for 241 frames, which is the highest resolution and frame count in the nuScenes dataset. However, the inference cost is currently substantial. Therefore, the primary numerical results in our paper do not utilize this maximum setting.
We have included some generated results in \appref{app:more vis} and on our project website for reference.
Future work may focus on further reducing the inference cost.

\section{Sequence Parallel Training}\label{app:seq_par}
\begin{figure}[h]
    \centering
    \includegraphics[width=0.75\linewidth]{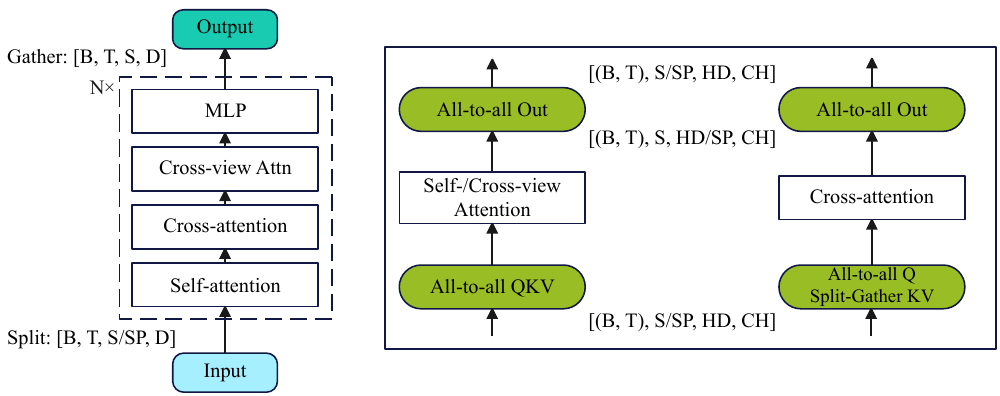}
    \caption{\textbf{Diagram for Sequence Parallel}.
    \textit{Left}: We split the spatial dimension before the first block and gather them after the last block.
    \textit{Right}: For each attention module, we use all-to-all communication, changing the splitting dimension to attention heads.
    B: batch; T: temporal dimension; S: spatial dimension; D: latent dimension; HD: number of heads; CH: per-head dimension; SP: sequence parallel size.
    }
    \label{fig:seq}
\end{figure}

Inspired by \citet{opensora}, we employ sequence parallelism to train DiT models with large sequence lengths. As illustrated in Figure \ref{fig:seq}, we partition each input across the spatial dimension onto different GPUs. Most operations can be executed within a single GPU; however, the attention blocks necessitate communication. On the right side of Figure \ref{fig:seq}, we demonstrate the communication process, where the full sequence is gathered, but the attention heads are distributed across different GPUs. This approach allows for peer-to-peer communication between GPUs while maintaining a roughly equal load.

Additionally, for VAE encoding and decoding, we partition based on batch size and the number of camera views, leveraging multiple GPUs to accelerate processing.

\section{More Details for Mixed Resolution and Frames Training}\label{app:mix}
\methodname is trained through a progressive training approach with variable length and resolution data configurations (see \secref{sec:bootstrap-variable}).
Consequently, our method of data mixing corresponds to the three training stages, as detailed in \tabref{tab:mix-detail}.

\input{tables/mix-detail}

Inspired by \cite{opensora}, to maximize the utilization of GPU resources, we employed a bucket-like approach to adjust the data composition. Specifically, each GPU process (or sequence parallel communication group) loads only one type of data to ensure alignment of the batch dimension.
Using the training time of the video format with the longest iteration time at batch size = 1 as a benchmark, we adjusted the batch sizes of other data formats so that each type runs at approximately the same speed. Notably, during stage 3 training, due to the limited number of full video clips, we repeat this type of data within an epoch. This ensures that different types of data have a similar magnitude of batch numbers within an epoch.

\section{Efficiency of Progressive Bootstrap Training}
The three-stage progressive training approach markedly improves model training efficiency relative to direct Stage 3 training. Table \ref{tab:speed} indicates that, over 4 days, for example, Stage 1 executes approximately 60 times more iterations than Stage 3, and Stage 2 achieves about 7 times more iterations. The progressive training is vital for controlled generative models, which require extensive iterations for effective convergence, as discussed in \secref{sec:bootstrap-variable}. The progressive strategy enables the rapid acquisition of high-quality video generation capabilities, utilizing faster iterations in the early stages to enhance convergence and expedite learning.

\input{tables/time}

\section{Video Generation Speed}
Table~\ref{tab:infer-cost} shows the breakdown of computation and inference costs, together with comparisons with others.
By adopting the sequence parallel, our inference speed is on par with the performance of NVIDIA's Cosmos-transfer1~\cite{nvidia2025cosmosdrivedreams}. We open-source our implementation and welcome future efforts on optimization.

\begin{table}[htb]
\centering
\begin{tabular}{@{}l|ccc|rr|r|c@{}}
\toprule
Method &
  resolution &
  \begin{tabular}[c]{@{}c@{}}\# \\ views\end{tabular}
  &
  \multicolumn{1}{l|}{
  \begin{tabular}[c]{@{}c@{}}\# \\ frames\end{tabular}
  } &
  \begin{tabular}[c]{@{}c@{}}Diff. Steps \\ (sec/it)\end{tabular} &
  \multicolumn{1}{c|}{\begin{tabular}[c]{@{}c@{}}Latent Dec. \\ (sec)\end{tabular}} &
  \multicolumn{1}{c|}{\begin{tabular}[c]{@{}c@{}}Total \\ (min)\end{tabular}} &
  Device \\ \midrule
  \multirow{4}{*}{
  \begin{tabular}[c]{@{}l@{}}\textit{MagicDrive-V2}\end{tabular}
  } &
  848x1600 &
  3 &
  193 &
  18.03 &
  82.83 &
  \multicolumn{1}{l|}{11.68} &
  \multirow{4}{*}{H20} \\
  &&&&& 248.24 (1 GPU) &\\
                                  & 848$\times$1600 & 6 & 241 & 53.74 & 103.36 & 28.92 &                            \\
                                  & 848$\times$1600 & 6 & 121 & 28.18 & 51.94  & 8.27  &                            \\ \midrule
\multirow{2}{*}{
\begin{tabular}[c]{@{}l@{}}Cosmos-transfer1~\cite{nvidia2025cosmosdrivedreams}\end{tabular}
} & 704$\times$1280 & $^{\dagger}$6 & 121 & 20.88 & 54     & 19.92 &  \multirow{2}{*}{\begin{tabular}[c]{@{}c@{}}A100-SXM4\end{tabular}} \\
                                  & 704$\times$1280 & 1 & 121 & 3.48  & 9      & 3.32  &                            \\ \bottomrule
\end{tabular}%
\caption{\textbf{Inference speed of \methodname}. Unless otherwise specified, we use 8 GPUs for testing. As compared with Cosmos-transfer1, our implementation of \methodname offers reasonable inference speeds. We also implemented a parallel decoding strategy for latent decoding, offering 3$\times$ speedup over 1 GPU.
Total time contains overhead from CPU offloading.
$^{\dagger}$There is no such a model; this row is estimated by $6\times$ the single-view time.}
\label{tab:infer-cost}
\end{table}

\section{More Experimental Details}\label{app:exp_setup}

The nuScenes~\cite{nuScenes} dataset includes 12Hz unannotated data and 2Hz annotated data. According to our experiments, high-frame-rate videos are more beneficial for generative model learning. Therefore, we follow \cite{magicdrive} and interpolated the 2Hz annotations to 12Hz annotations with ASAP~\cite{wang2023we}. Although the interpolation results are not entirely accurate, they do not affect the training for video generation.
The Waymo Open Dataset~\cite{Sun_2020_CVPR} includes 10Hz annotated data. We follow the official splitting, which has 798 clips for training and 202 clips for validation.
The original dataset contains 5 views. However, the dimensions of the left and right side views are smaller than those of the three front views, and their field of view is limited. Therefore, we only retain three front views for training and validation.

\noindent\textbf{Semantic Classes for Generation}.
We follow ~\cite{magicdrive} in data setup on nuScenes.
Specifically, for objects, ten categories include car, bus, truck, trailer, motorcycle, bicycle, construction vehicle, pedestrian, barrier, and traffic cone.
For the road map, eight categories include drivable area, pedestrian crossing, walkway, stop line, car parking area, road divider, lane divider, and roadblock.
For Waymo, the semantics for objects include pedestrian, car, and cyclist, while the semantics for road maps include drivable area, crosswalk, road line (yellow), and road line (white).

\section{More Training Details}\label{app:exp_train}

\noindent\textbf{Optimization}.
We train our diffusion models using Adam optimizer and a constant learning rate at $8e^{-5}$ with a 3000-step linear warm-up in the last two stages. We primarily use 32 NVIDA A800 GPUs (80G) for training. Our model can also be trained with Ascend 910B (64G). The batch size for each stage is set according to the iteration speed, following the bucket strategy as \cite{opensora}. For example, in stage 2, we set the batch size for 33$\times$424$\times$800 to 1, which takes about 30s/it. Then we set the batch size to other video types to achieve about 30s/it. This strategy can ensure the load balance between different GPU processes.

\noindent\textbf{Inference}.
By default, images/videos are sampled using Rectified Flow~\citep{esser2024scaling} with 30 steps and the classifier-free-guidance (CFG) scale at $2.0$.
To support CFG, we randomly drop different conditions at a rate of 15\%, including embeddings for text, camera, ego trajectory, and boxes. We follow \citet{magicdrive} to use $\{\mb{0}\}$ as the \textit{null} condition for maps in CFG inference.
When inferring high-resolution long videos, we also use sequence parallel (\appref{app:seq_par}) to fit in the limited memory of a single GPU.

\section{Human Evaluation for Multi-frame \& Multi-view Consistency}\label{app:human-eval}
\begin{figure}[ht]
    \centering
    \includegraphics[width=0.9\linewidth]{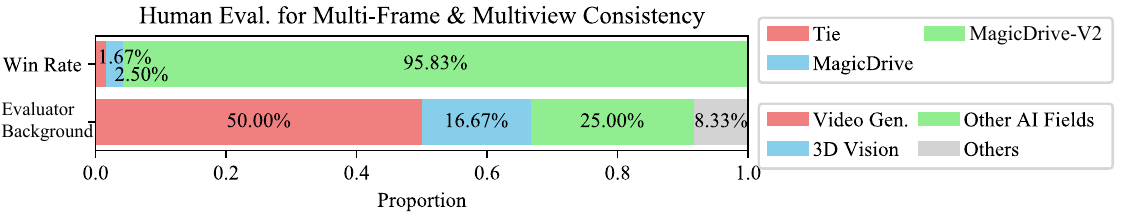}
    \caption{Experts evaluated content consistency against \citet{magicdrive}, showing \methodname generates videos with superior consistency.}
    \label{fig:human-eval}
\end{figure}

Evaluating multi-frame and multi-view consistency in video generation has long been a challenging issue~\cite{agarwal2025cosmos}, as the academic community lacks a unified standard to accurately assess such consistency.
To address this, we employed human evaluation to measure these two aspects of consistency.
Specifically, we invited participants with diverse backgrounds to compare videos generated by \methodname and \citet{magicdrive} under identical conditions. 
Participants were asked to select the video with better consistency, and the winning probability of each model was statistically analyzed.
As illustrated in \figref{fig:human-eval}, \methodname demonstrated significantly superior consistency, indicating a substantial improvement over \citet{magicdrive}.

\begin{figure}[htb]
    \centering
    \vspace{-0.4cm}
    \includegraphics[width=0.45\linewidth]{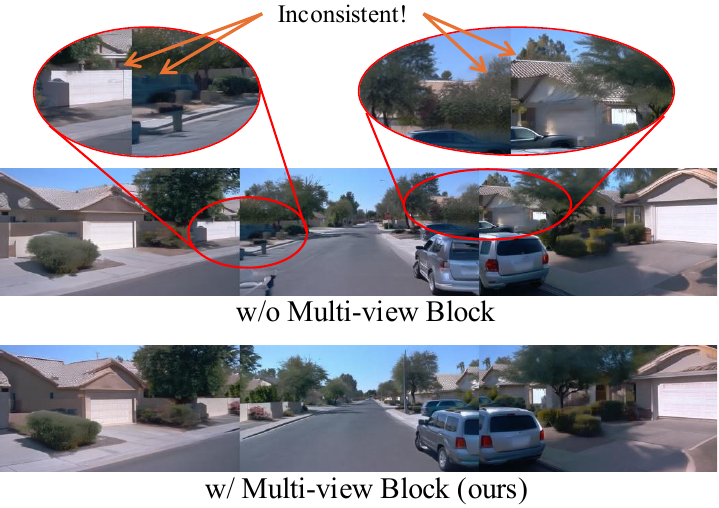}
    \vspace{-0.35cm}
    \caption{\textbf{Ablation on the MVDiT block}. MVDiT block is the key to
    enabling multi-view consistency in driving video generation.}
    \label{fig:mv-ablation}
    \vspace{-0.1cm}
\end{figure}

Besides, \figref{fig:mv-ablation} provides an ablation comparison between with and without the MVDiT block, showing that it is crucial to include such a block for multi-view consistency.

\section{Human Evaluation on Text Control}
\begin{figure}[htb]
    \centering
    \includegraphics[width=0.45\textwidth]{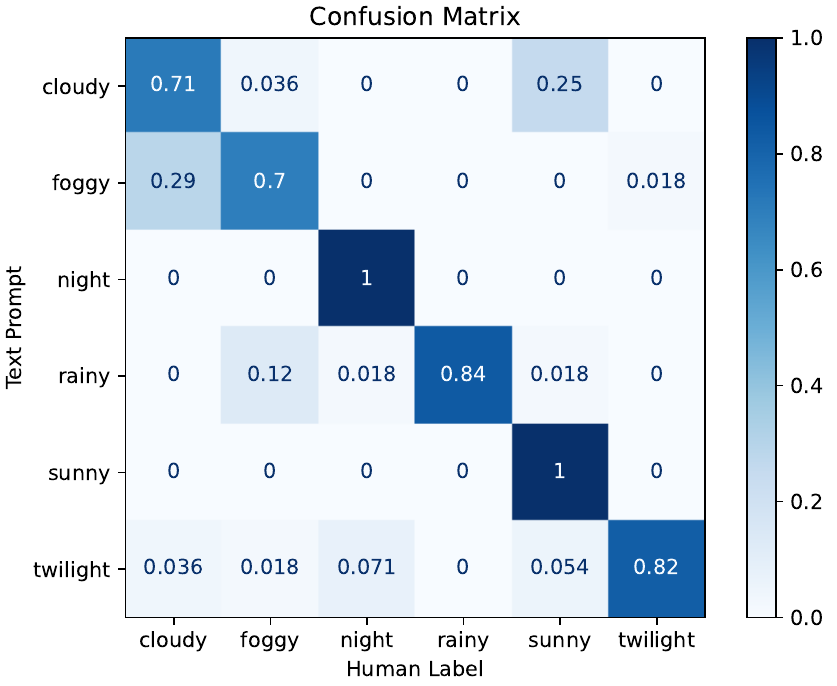}
    \caption{\textbf{Confusion matrix to show the text control ability in \methodname}. \methodname supports the control of 6+ weather conditions.}
    \label{fig:text-right}
\end{figure}

We validate text control by generating videos under six weather conditions and asking humans to judge alignment with text prompts. The confusion matrix in \figref{fig:text-right} shows high recognition accuracy ($>70\%$), confirming effective text control.

\input{tables/vae}

\section{More Comparison among VAEs}\label{app:vae_comp}
To quantitatively compare the performance of the VAE, we randomly selected two 6-view videos from the nuScenes dataset and used the PSNR metric to evaluate the VAE's reconstruction ability. \tabref{tab:vae} presents the results, averaged across the six views. From these results, we observe that CogVAE~\cite{yang2024cogvideox} demonstrates the best reconstruction ability, even surpassing the 2D VAE~\cite{rombach2021highresolution}.
Comparing the results from different settings, we find that the current 3D VAEs exhibit good generalization ability for long videos, primarily due to the window-based downsampling techniques~\cite{yang2024cogvideox,opensora}.
Additionally, we observe that high-resolution content retains a relatively high PSNR after VAE reconstruction, indicating that the current VAEs are more favorable for high-resolution data.
This observation also supports our motivation for high-resolution generation.

\section{Reason for Using CogVAE without the Pre-trained Diffusion Model}\label{app:vae}

Informed by the work of MagicDrive~\cite{magicdrive} and others~\cite{gao2024vista}, fine-tuning from a well-performing diffusion model can effectively accelerate model convergence.
Consequently, in the initial implementation of the DiT architecture, we experimented with Open-Sora 1.2~\cite{opensora}’s VAE and diffusion models. However, the results were suboptimal, with image generation and video controllability falling short of MagicDrive's performance. We attribute this primarily to the limited generalization capability of text-to-video diffusion and, more critically, to the inadequate reconstruction ability of the VAE.

We conducted a comparative analysis of VAEs, as detailed in \secref{sec:ablation} and \appref{app:vae_comp}, and found CogVAE~\cite{yang2024cogvideox} to perform well. Given that the VAE determines the upper limit of generation quality, we opted to use CogVAE for video encoding. Notably, CogVideoX~\cite{yang2024cogvideox} employs a novel DiT structure, where each layer's latent space integrates both video and text condition information. This approach may complicate the design of geometry-related conditions. Furthermore, CogVideoX was not trained in a driving scenario. To eliminate these potential confounding factors, we decided to train the diffusion model from scratch using CogVAE. This strategy allows us to move beyond the constraints of pre-trained models, enabling more flexible modifications to the model architecture to achieve multi-view consistency and spatiotemporal encoding of geometry conditions.

Our experience directly demonstrates that high-resolution, long street-view video generation does not necessarily require pre-trained image-text or video-text models. Even so, this is beyond the primary focus of our paper, and we leave related questions for future work.

\section{Single Inference v.s. Rollout Inference}\label{app:vsrollout}

\begin{figure}[h]
\vspace{-0.4cm}

    \centering
    \begin{subfigure}[h]{\textwidth}
        \centering
        \includegraphics[width=0.95\textwidth]{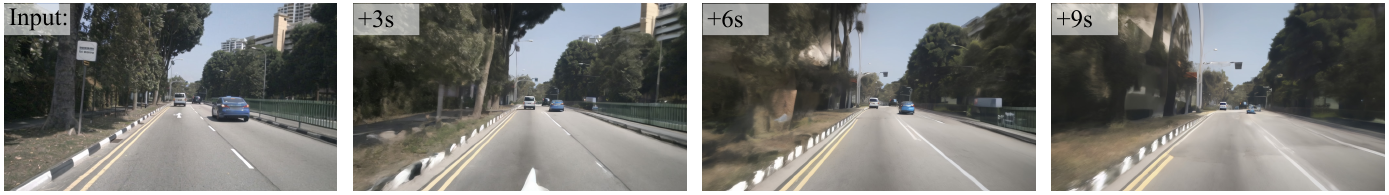}
        \caption{\textbf{Generation from Vista}. It takes the first frame as input and generates the following (only supporting the front view).}
        \label{fig:subfig1}
    \end{subfigure}

    \begin{subfigure}[b]{\textwidth}
        \centering
        \includegraphics[width=0.95\textwidth]{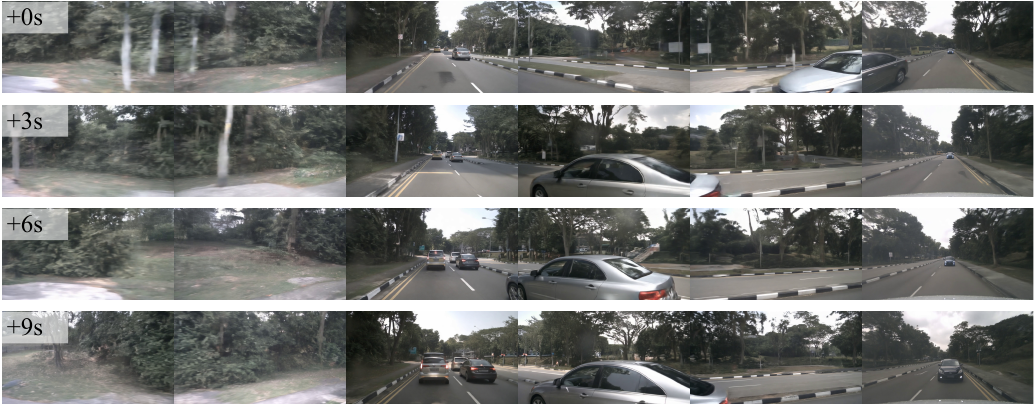}
        \caption{\textbf{Generation from \methodname}. We take conditions as inputs and generate the full video (only show the first 9s for comparison).}
        \label{fig:subfig2}
    \end{subfigure}
    \vspace{-0.5cm}
    \caption{\textbf{Comparison between Rollout for Long Videos (Vista~\cite{gao2024vista}) and Single Inference (our \methodname)}. Although rollout can handle long videos, the quality is significantly degraded. In contrast, our extrapolation maintains high quality in long video generation.}
    \label{fig:mainfig}
    \vspace{-0.5cm}
\end{figure}

To achieve long video generation, previous work typically employs a method of future frame prediction combined with rollout. This involves, after the $n$-th inference, taking the last $l$ frames from this inference as the first $l$ frames for the $n+1$-th inference, thus enabling long video generation. However, since the model does not directly capture long-term dependencies and accumulates errors with each inference, such rollouts often fail to support sufficiently long videos. Among rollout methods, Vista~\cite{gao2024vista} currently achieves relatively good results. We compared a 9-second video generated by performing 4 rollouts with Vista (the paper claims it can support 6 rollouts) to a 9-second segment produced by our method, \methodname. It is evident that our method maintains consistent video quality over long sequences, whereas Vista's results show a noticeable decline. Therefore, we believe that the hybrid training and length extrapolation approach adopted by \methodname can achieve higher quality in long video generation.

\section{More Visualization}\label{app:more vis}
\begin{figure}[p]
    \centering
    \includegraphics[height=8.575in]{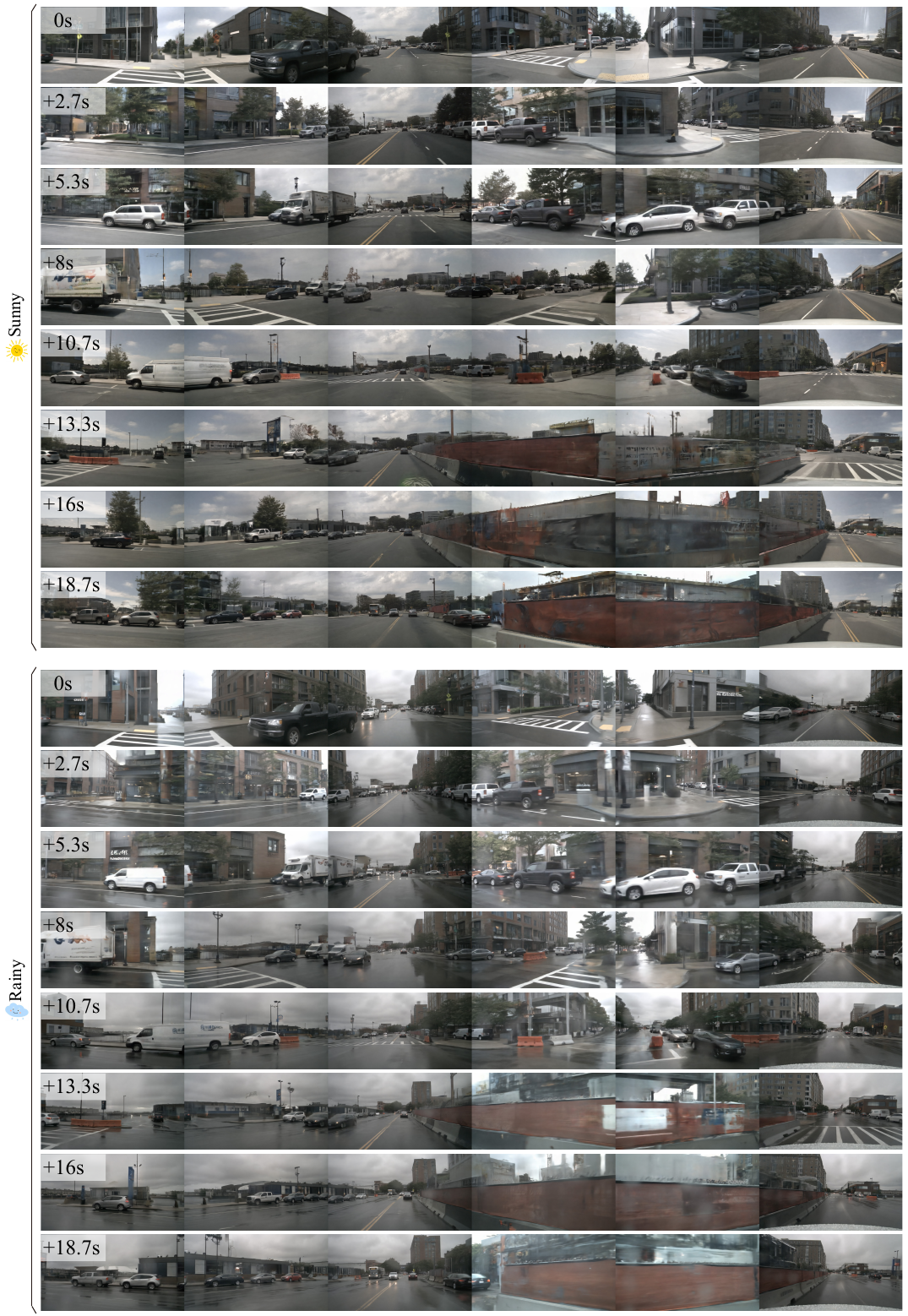}
    \vspace{-0.3cm}
    \caption{We show some frames from the generated $6\times848\times 1600\times 241$ videos with the same scene configuration (\ie, boxes, maps, cameras, and ego trajectory) but under different weather conditions. Conditions are from the nuScenes validation set.}
    \label{fig:more-exp1}
    \vspace{-0.3cm}
\end{figure}

\begin{figure}[p]
    \centering
    \includegraphics[height=8.575in]{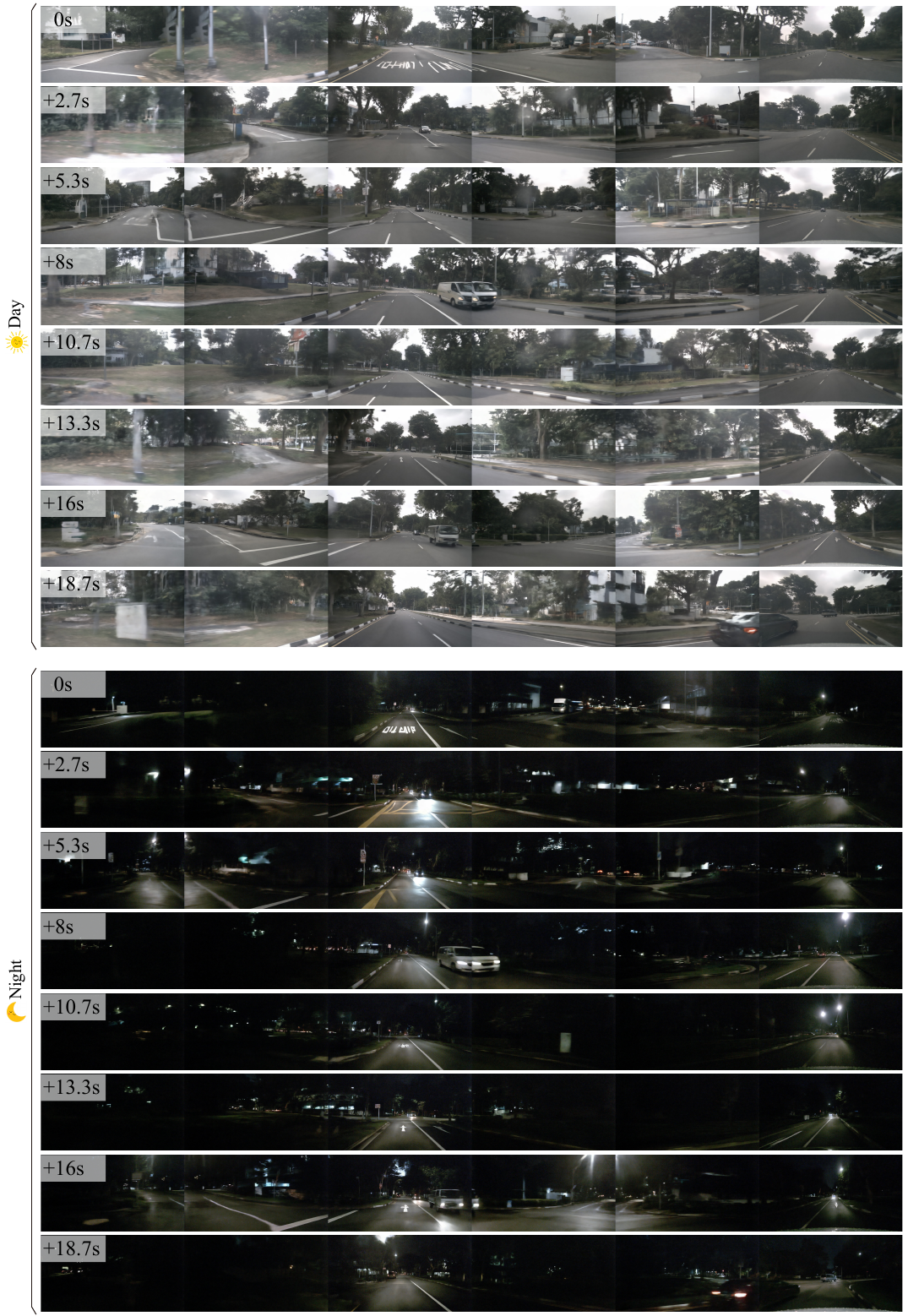}
    \vspace{-0.3cm}
    \caption{We show some frames from the generated $6\times848\times 1600\times 241$ videos with the same scene configuration (\ie, boxes, maps, cameras, and ego trajectory) but under different time-of-day conditions. Conditions are from the nuScenes validation set.}
    \label{fig:more-exp2}
    \vspace{-0.3cm}
\end{figure}

As said in the ``Note'', \methodname is capable of generating $6\times848\times1600\times241$ videos (20s at 12 fps). We include more generated samples in \figref{fig:more-exp1}-\ref{fig:more-exp2}.
Please see the videos on our project page.

%% file: tables/mix-detail.tex
\begin{table}[ht]
\centering
\begin{tabular}{l|l|l|c|c}
\toprule
Stage & Resolution & Frame(s) & Sequence Parallel & Training Step \\ \midrule
Stage 1 & 224$\times$400 & Img & - & 80000 \\ \midrule
\multirow{2}{*}{Stage 2} & 224$\times$400 & Img, 9, 17, 33, 65 & \multirow{2}{*}{-} & \multirow{2}{*}{40000} \\
 & 424$\times$800 & Img, 9, 17, 33 &  &  \\ \midrule
\multirow{3}{*}{Stage 3} & 224$\times$400 & Img, 17, full & \multirow{3}{*}{4} & \multirow{3}{*}{30000} \\
 & 424$\times$800 & Img, 17, 33, 65, 129 &  &  \\
 & 848$\times$1600 & Img, 9, 17, 33 &  &  \\ \bottomrule
\end{tabular}
\caption{\textbf{Configuration for Variable Length and Resolution Training}.
The mixing configuration aligns with our progressive bootstrap training with 3 stages, from low-resolution images to high-resolution long videos.
}
\label{tab:mix-detail}
\end{table}

%% file: tables/time.tex
\begin{table}[t]
\centering
\begin{tabular}{l|r|c}
\toprule
Stages & Seconds/Iter. & Iter. for 4 days \\ \midrule
stage 1 & 4.32 & 80k \\
stage 2 & 39.84 & 8.7k \\
stage 3 & $^{*}$264.96 & 1.3k \\ \bottomrule
\end{tabular}
\vspace{-0.2cm}
\caption{
\textbf{Speed for Each Training Stage of \methodname}, measured on NVIDA A800 GPUs.
Over a 4-day period (for example), Stage 1 training yields nearly 60 times more iterations than Stage 3, and Stage 2 offers about 7 times more.
$^{*}$This value is calculated by multiplication with sequence parallel (SP) size (in practice, we use SP size of 4 for the stage 3, with 66.24s/it). 
}
\label{tab:speed}
\vspace{-0.3cm}
\end{table}

%% file: tables/vae.tex
\begin{table}[ht]
\centering
\begin{tabular}{l|l|ccc}
\toprule
Resolution & Model & Image & 17 fr. & 33/34 fr. \\ \midrule
\multirow{3}{*}{224$\times$400} & CogVAE & \textbf{34.4261} & \textbf{31.0900} & \textbf{30.5986} \\
 & Open-Sora & 30.4127 & 27.9238 & 27.5245 \\
 & SD VAE & 27.7131 & 27.7593 & 27.9404 \\ \midrule
\multirow{3}{*}{424$\times$800} & CogVAE & \textbf{38.4786} & \textbf{33.5852} & \textbf{32.9202} \\
 & Open-Sora & 33.6114 & 30.2779 & 29.8426 \\
 & SD VAE & 30.9704 & 31.0789 & 31.3408 \\ \midrule
\multirow{3}{*}{848$\times$1600} & CogVAE & \textbf{41.5023} & \textbf{36.0011} & \textbf{35.1049} \\
 & Open-Sora & 37.0590 & 33.2856 & 32.8690 \\
 & SD VAE & 37.0504 & 33.2846 & 32.8680 \\ \bottomrule
\end{tabular}
\vspace{-0.2cm}
\caption{\textbf{VAE Comparison for Street Views}.
CogVAE~\cite{yang2024cogvideox} and Open-Sora~\cite{opensora} (1.2) are 3D VAEs; SD VAE~\cite{rombach2021highresolution} is 2D VAE, which is also widely adopted by previous street view generation (\eg, \cite{magicdrive}).
Results are PSNRs calculated through videos from the nuScenes validation set.
\methodname adopts CogVAE.
}
\label{tab:vae}
\vspace{-0.5cm}
\end{table}